\documentclass{article}
\usepackage{acra}

\usepackage{graphicx}
\usepackage{amsmath,amssymb} 
\usepackage{color}
\usepackage{array}
\usepackage{outlines}
\usepackage{caption} 
\usepackage{algorithmic}
\usepackage{algorithm}
\usepackage{hyperref}
\usepackage{cleveref}
\usepackage{wrapfig}
\usepackage{subfigure}
\usepackage{booktabs}
\usepackage{balance}
\usepackage{tabularx}

\newcommand{\degree}{$^{\circ}$}

\begin{document}

\title{Accessible Torque Bandwidth of a Series Elastic Actuator Considering the Thermodynamic Limitations}

\author{Bhanuka Silva\textsuperscript{1} and Navinda Kottege\textsuperscript{2} \\
\textsuperscript{1}Department of Electronic and Telecommunication Engineering, University of Moratuwa, Sri Lanka\\
\textsuperscript{2}Robotics and Autonomous Systems Group, CSIRO, Pullenvale, QLD 4069, Australia}

\maketitle
\thispagestyle{empty}
\pagestyle{empty}

\begin{abstract}
Within the scope of the paper, electromechanical and thermodynamic models are derived for a series elastic actuator and open loop and closed loop torque bandwidth parameters are analysed considering the thermodynamic behaviour of the actuator. It was observed that the closed loop torque bandwidth of the electromechanical subsystem of the actuator was not accessible in the entire torque reference amplitude range due to thermodynamic limitations. Therefore, a stator winding temperature estimation based adaptive controller is utilised and analysed to improve the accessibility of the controller based torque bandwidth. This paper implements the methodology on a HEBI Robotics X5-9 actuator as a case study.
\end{abstract}

\section{Introduction}

In many robotic applications, it is required to control robot-environment interaction forces. For this purpose, it is essential to have high performance actuators providing the ability of controlling applied torque to the joints of mechanisms.

In series elastic actuators (SEAs), an elastic element is placed between the load side and the gear reduced electric motor side. Many previous works like \cite{R6}, \cite{R2} and \cite{R5} discuss the inherent pros and cons of such actuators. The deflection in the elastic element produces the required amount of torque to the load side of the actuator, therefore, force control based strategies can be converted to position control based strategies of the elastic element's deflection which is one of the main advantage of SEA based control systems \cite{R6}. A careful parameter identification of the SEA electromechanical subsystem allows the utilisation of model based feedback and disturbance rejection controllers for the actuator output's robust torque tracking accuracy. 

Neglecting thermal limitation analysis in torque control applications has a risk factor of driving the internal motor to extreme stator winding temperatures resulting in irreversible internal demagnetisation damages \cite{R1}. Torque control algorithms and thermodynamic models are often treated as separate branches for an actuator, therefore, the literature rarely adopts the analysis of closed loop torque bandwidth of an actuator considering the thermodynamic performance limitations. This paper assesses the methodology in modelling electromechanical and thermodynamic subsystems and conducting experiments to bridge the gap. 

\begin{figure}[t!]
	\centering
	\includegraphics[width=0.49\textwidth]{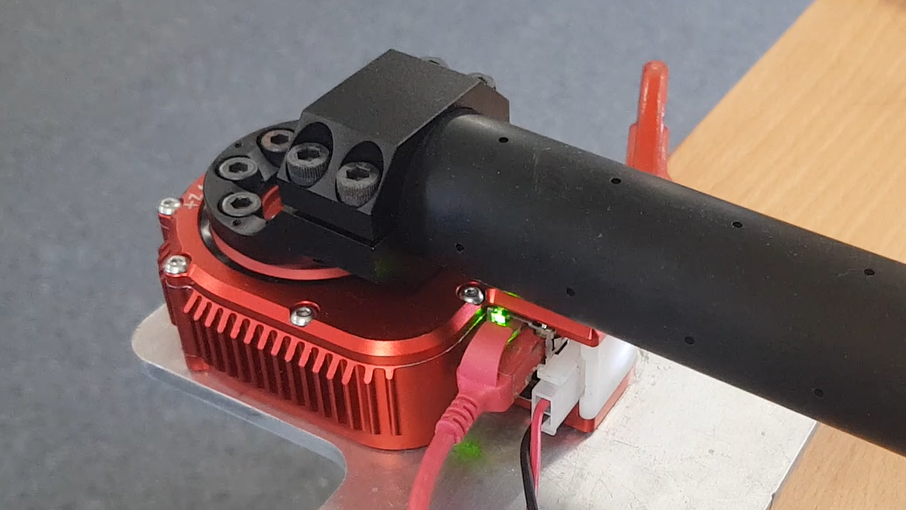}
	\caption{X5-9 series elastic actuator from HEBI Robotics. Casing is mounted to a fixed metal base while the actuator output is fixed for an experiment.}
    \label{fig:hebi}
\end{figure}

\subsection{Related work}

For the purpose of modelling of a SEA, \cite{R5} exploit a two-mass model which includes the interaction of the load side dynamics that is modelled using a simple inertia and a damping coefficient. \cite{R2} add an additional layer of impedance interaction for modelling any interaction with the environment. Therefore, the two mass model will effectively characterise most of the linear dynamics of the actuator. They improve force control performance using a disturbance observer (DOB) and a feedforward controller and explore the robust stability of the system against changes in environmental parameters. \cite{R7}  discuss a novel method to use an adaptive disturbance observer to estimate the parameters of the nominal plant inverse model which in turn minimises the errors of the DOB. \cite{R9} and \cite{R10} further discuss the observer design and \cite{R4}  reveal that an open loop DOB where the observer is applied only to the plant model and a closed loop DOB where the observer is applied to the plant as well as the controller would behave in a similar manner in enforcing the nominal plant model. The former implementation is associated in this paper. 

For a model based control system design, careful parameter identification of SEA is important. Brushless direct current (BLDC) motors, due to their high efficiency, less maintenance, wide range of operating speed, high power density, small size and good torque-speed characteristics are preferred over brushed DC motors. However, the dynamic behaviour of BLDC motors is non-linear \cite{R11}. The torque ripple effect caused by multiple phases and magnetic fields is considered one of the main reasons for such non-linearities \cite{R8}. A combination of feedback linearisation strategy and an observer is often used to enforce the nominal plant model by compensating for such effects.

\cite{R1}  assess the physically feasible torque control bandwidth of an integrated toque controllable actuator and the controller independent physical upper bound for the bandwidth of the actuator when the output of the actuator is fixed as well as varied in inertia. A safe operation time is computed for each torque-frequency condition by looking at the thermal actuator properties. For the inclusion of thermal limitations, thermodynamic modelling of an actuator is important and lumped parameter modelling technique is argued to have good accuracy in correctly predicting the thermal characteristics in a wide range of operating conditions \cite{R12}, \cite{R13} and \cite{R14}.

The organisation of the paper is as follows. Section \ref{methodology} discusses the methodology of SEA modelling and controller implementation. Section \ref{results} emphasises on the torque tracking results that are obtained in conjunction with thermodynamic behaviour. The paper concludes in section \ref{conclusion} discussing the effectiveness of the novel methodology in maximising the accessible torque bandwidth.

\section{Methodology} 
\label{methodology}

This section discusses the electrical, mechanical and thermodynamic subsystems of a SEA emphasising the modelling approximations and parameter identification methodologies. A brief theoretical analysis on DOBs and model based controller designing are also discussed in this section for applied torque's robust tracking accuracy.

\begin{figure}[t]
	\centering
	\includegraphics[width=8.5cm]{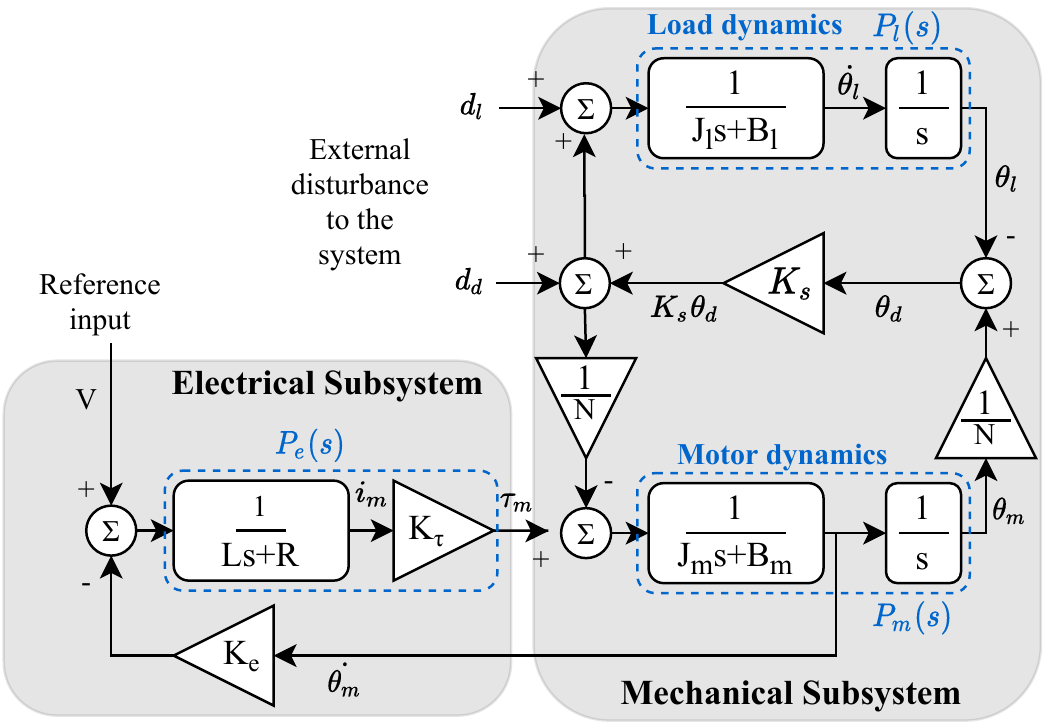} 
	\caption[Optional caption]{Linear model of the SEA.}
    \label{figure:SEAmodel}
\end{figure}

\subsection{Electrical subsystem}

The SEA input voltage is regulated to the nominal value and an internal controller regulates the voltage to the terminals of the BLDC motor to generate electromagnetic torque $\tau_m(t)$. Readings of three hall sensors at 120\degree\,offset are used allowing six different control voltage combinations for the three phases.

Linear dynamics of the electrical subsystem are characterised by effective winding resistance $R$ and winding inductance $L$. The following equations represent the dynamics of the electrical subsystem of the actuator. $K_e$ and $K_{\tau}$ represent the back electromotive force constant and motor torque constant respectively. $V(t)$ is the applied winding voltage and $i_m(t)$ is the winding current of the linear model. $\delta(t)$ is the Dirac delta function. 

\begin{equation}
	\label{second electrical equation}
	V(t)-K_{e}\dot{\theta}_{m}(t)=Ri_m(t)+L(\dot{i_m}(t)+i_m(0)\delta(t))
\end{equation}

\begin{equation}
	\label{third electrical equation}
	\tau_m(t)=K_{\tau}i_m(t)
\end{equation}

\subsection{Mechanical subsystem}

Time domain linear equation given in equation \ref{first mech equation} characterises the mechanical subsystem of the BLDC motor. The rotor inertia is $J_m$ and the viscous friction is $B_m$. The effects of cogging torque, frictional torque and coulomb frictional torque are neglected as non linearities to be compensated by the DOB. $K_s$ represents the torque constant of the elastic element and the generated torque is considered to have a static characteristic as in equation \ref{second mech equation}. Deflection of the elastic element is represented by $\theta_d$ and the BLDC motor angle by $\theta_m$. Inertia, damping  and frictional torques of the gearbox are neglected for the simplicity of the actuator model and has a gear reduction of $N$ to $1$. 

\begin{equation}
	\label{first mech equation}
	\tau_m(t)-\frac{K_s\theta_d(t)}{N}= J_m(\ddot{\theta}_m(t)+\dot{\theta}_m(0)\delta(t))+B_m\dot{\theta}_m(t)
\end{equation}

\begin{equation}
	\label{second mech equation}
	\tau_{spring}=K_s\theta_d
\end{equation}

Linear dynamics of the load $A(s)$ are parameterised by the load inertia $J_l$ and the damping coefficient $B_l$. In general, load inertia can be represented by a time varying functional depending on the application of the actuator. The stiffness of the load side is assumed to be zero.

\begin{equation}
	\label{third mech equation}
	A(s)=J_ls+B_l
\end{equation}

Figure \ref{figure:SEAmodel} depicts the linear model of the SEA. Note that Laplace domain notation `s' is removed for simplicity. Transfer functions represented by $P_e(s)$, $P_m(s)$ and $P_l(s)$ are shown in the same figure. For the electrical subsystem;

\begin{equation}
\label{matrix part 2}
    \begin{bmatrix}
         \tau_m\\
    \end{bmatrix}=
    \begin{bmatrix}
         P_e & -sK_{e}P_e\\
    \end{bmatrix}
    \begin{bmatrix}
         V\\
         \theta_m\\
    \end{bmatrix}
\end{equation}

For the mechanical subsystem;

\scriptsize{
\begin{equation}
\label{matrix part 1}
    {\small
    \begin{bmatrix}
     \theta_d\\
     \theta_l\\
     \theta_m \\
    \end{bmatrix}=
    \begin{bmatrix}
         \frac{P_mN^{-1}}{D} & -\frac{N^{-2}P_m+P_l}{D} & -\frac{P_l}{D}\\
         \frac{K_sP_lP_mN^{-1}}{D} & \frac{P_l}{D} & \frac{P_l(1+K_sN^{-2}P_m)}{D}\\
         \frac{P_m(1+K_sP_l)}{D} & -\frac{N^{-1}P_m}{D} & \frac{K_sN^{-1}P_mP_l}{D}\\
    \end{bmatrix}
    \begin{bmatrix}
         \tau_m\\
         d_d\\
         d_l\\
    \end{bmatrix} 
    }
\end{equation}
}

\normalsize{
Here $D(s)= 1+K_sN^{-2}P_m(s)+K_sP_l(s)$.
}

\subsection{System identification of the electromechanical subsystem}

An open loop frequency sweep signal is fed while the output of the actuator is fixed simulating an infinite load side inertia ($J_l\rightarrow\infty$). The system transfer function $\frac{\theta_d(s)}{V(s)}$ obtained from equations \ref{matrix part 1} and \ref{matrix part 2} then reduces to; 

\begin{equation}
\label{output locked tf}
	\frac{\theta_d(s)}{V(s)}=\frac{\frac{K_{\tau}}{N}}{A_3s^3+A_2s^2+A_1s+A_0}
\end{equation}

Denominator coefficients are simplified in the equation \ref{output locked tf} and stand for; $A_0=\frac{K_sR}{N^2}$,  $A_1=K_eK_{\tau}+\frac{K_sL}{N^2}+B_mR$, $A_2=B_mL+J_mR$ and $A_3=J_mL$. Therefore, the transfer function identification involves in a third order model fit and experimental data are obtained varying the amplitude of the commanded PWM reference signal. However, for each amplitude value, there can be mismatches in the model fit results and it implies the presence of system non-linearities as well as time variant behaviours. A nominal plant model $P_{n}$ is derived from the results such that it minimises the mismatches. Uncertainty estimation of the model fit results with respect to the selected nominal plant model can be obtained from equation \ref{uncertainity eqn}. $G_{x}(s)$ represents the experimental model fit for the reference PWM amplitude value $x$.

\begin{equation}
	\label{uncertainity eqn}
	\Delta_x(s)=\frac{G_{x}(s)-G_{nominal}(s)}{G_{nominal}(s)}
\end{equation}

\subsection{Thermodynamics}
\label{thermodynamics}

\begin{figure}[t]
	\centering
	\includegraphics[height=6cm]{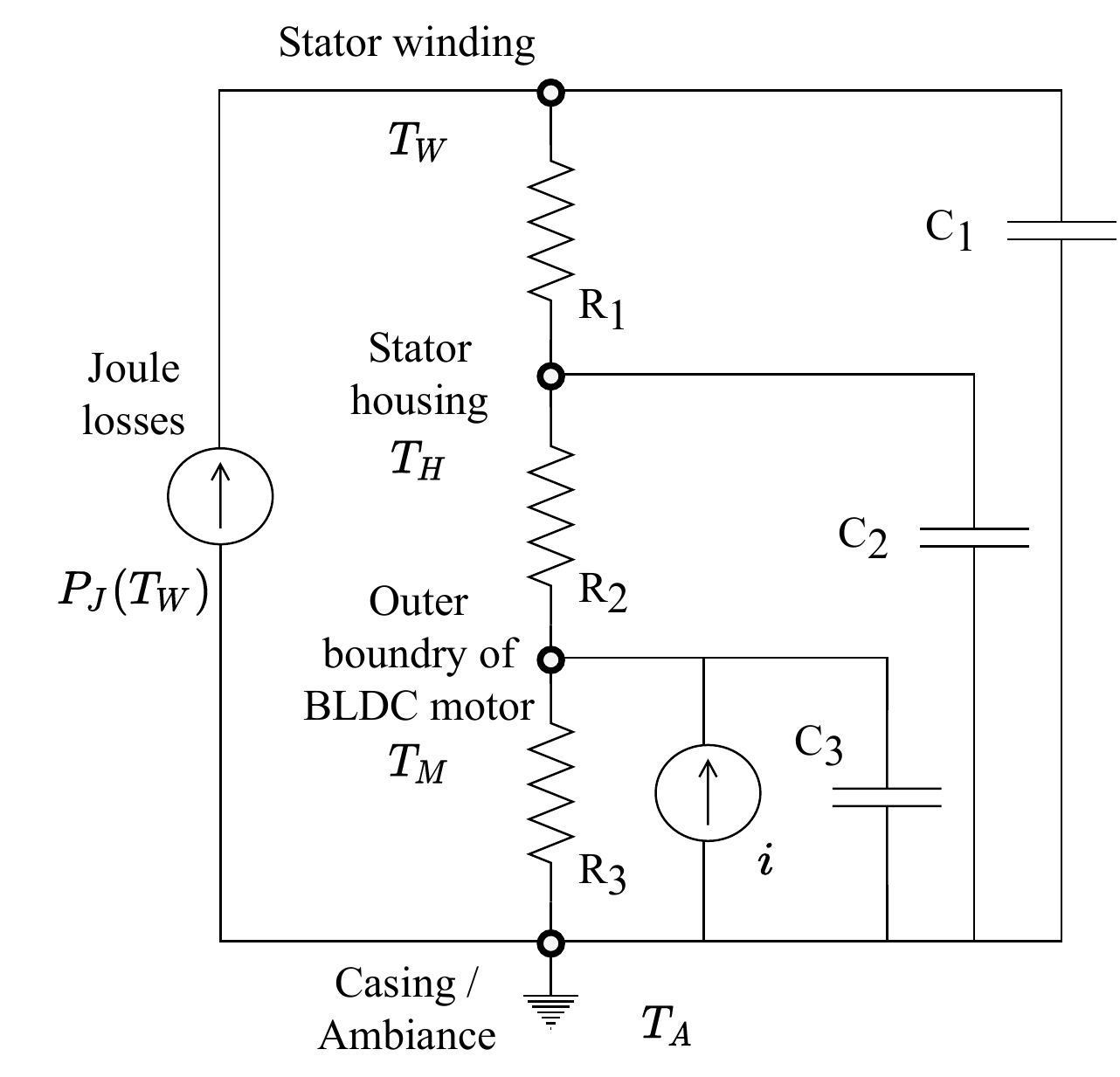} 
	\caption[Optional caption]{Four body electro analogous circuit diagram.}
    \label{figure:temp analogy}
\end{figure}

Heat energy flow per unit time within the thermodynamic subsystem of the actuator is modelled using the analogy of an equivalent electrical circuit. The electrical currents in the Figure \ref{figure:temp analogy} represent the power flow as a result of the heat flux transfer from the identified thermal bodies of the actuator. A five body electro-analogous network consisting motor stator winding, motor stator housing, outside of the motor, the controller electronics and the metallic case of the enclosed actuator could be observed. The network can be further simplified to a four body analogy with the assumption that the casing is equal to the ambiance as long as it is fixed to a large metal base consisting an approximated infinite heat capacity in it. The ambiance $T_A$ is controlled to be 25\degree C. Actuator's electronic circuit board temperature sensor in general reads a higher temperature ($T_B$) than the ambiance of 25\degree C due to solid-state electronic heat flux, therefore, the resultant power flow is modelled using a current source $i$ in the model.

In Figure \ref{figure:temp analogy}, thermal resistances $R_{1}$ to $R_{3}$ characterise the heat dissipation between the different bodies inside the actuator. $C_{1}$, $C_{2}$ and $C_{3}$ represent the heat capacity constants. $R_{1}C_{1}$ to $R_{3}C_{3}$ characterise the thermal time constants $\tau_1$ to $\tau_3$. $T_W$, $T_H$ and $T_M$ stands respectively for the instantaneous temperatures of the stator winding, stator housing and BLDC motor proximity at a given time.  

The power loss due to the stator winding current is called Joule power loss and is the significant source of heat flux of the model. It can be governed by the following equation where the electrical winding resistance is also temperature dependent which in turn cannot be neglected in thermal behaviour. $R_{A}$ is the electrical winding resistance at 25\degree C and $i_m$ is the stator winding current at a given time. $\alpha_{cu}$ represents the temperature coefficient of resistance.

\begin{equation}
	\label{node zero thermal equation}
	P_J(T_w)=R_{A}(1+\alpha_{cu}(T_W-T_A)){i_m}^2
\end{equation}

The differential equations \ref{node one thermal equation}, \ref{node two thermal equation} and  \ref{node three thermal equation} are obtained from the law of conservation of charge entering and leaving a junction in equivalent electrical circuit model of Figure \ref{figure:temp analogy}.

\begin{multline}
	\label{node one thermal equation}
	R_A(1 + \alpha_{cu}.(T_W-T_A)){i_m}^2=\\
	\frac{T_W-T_H}{R_{1}} + C_{1}\frac{d}{dt}(T_W-T_A)
\end{multline}

\begin{equation}
	\label{node two thermal equation}
	\frac{T_W-T_H}{R_1} = \frac{T_H-T_M}{R_2} + C_2\frac{d}{dt}(T_H-T_A)
\end{equation}

\begin{equation}
	\label{node three thermal equation}
	\frac{T_H-T_M}{R_2} + i = \frac{T_M-T_A}{R_3} + C_3\frac{d}{dt}(T_M-T_A)
\end{equation}

\begin{figure}[H]
	\centering
	\includegraphics[height=5.5cm]{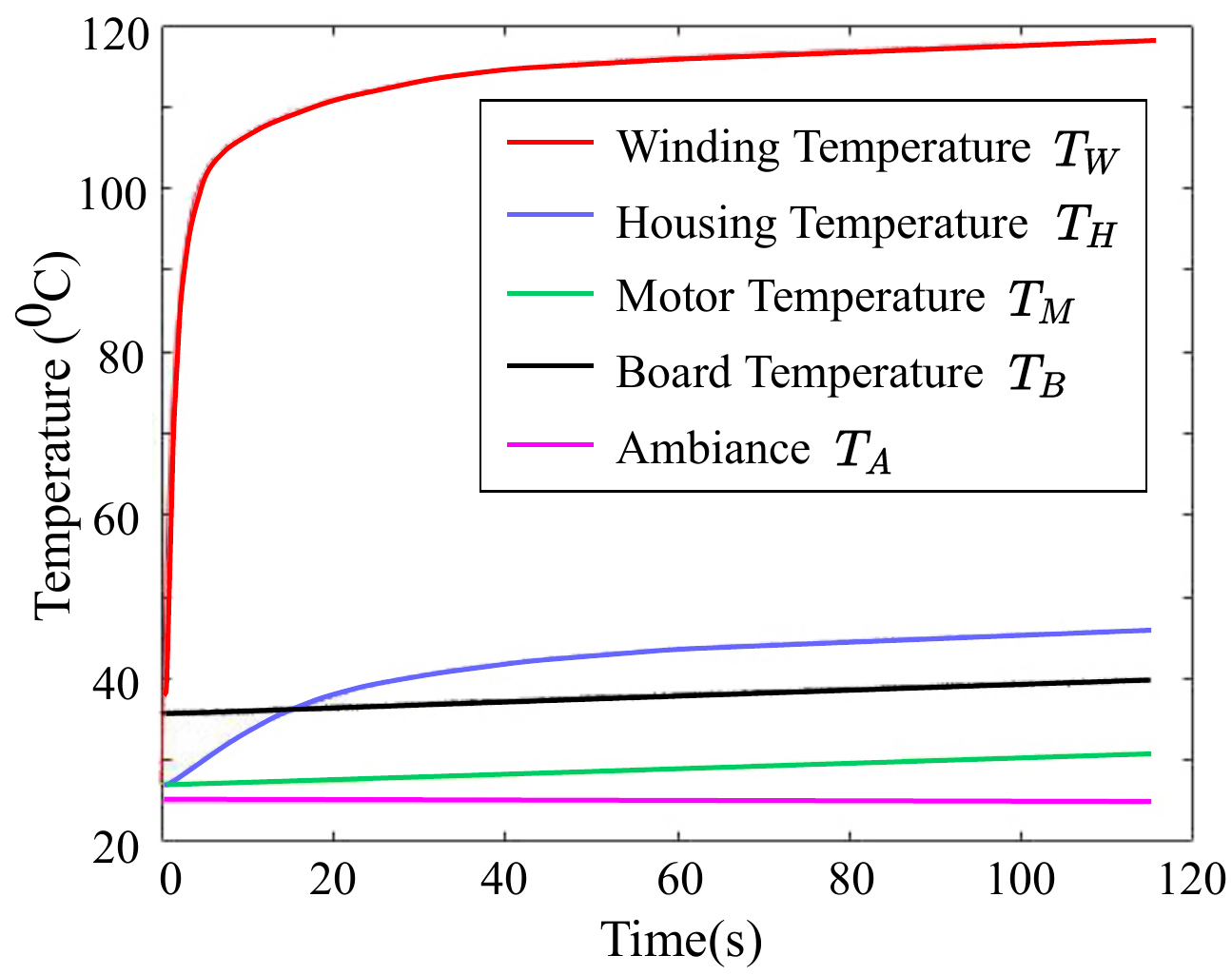} 
	\caption[Optional caption]{General four body temperature variation observation. }
    \label{figure:temp variation}
\end{figure}

Temperature variation of the thermal bodies is shown in Figure \ref{figure:temp variation} for a test case with the output fixed and commanded with an open loop PWM value resulting in a scaled step input of winding current. 

According to the temperature curves in Figure \ref{figure:temp variation}, $T_W$ has the smallest rise time compared to $T_H$ and $T_M$ and can have a higher probability in reaching the operative ceiling defined by the manufacturer. The maximum permissible winding temperature $T_{MAX}$ is 130\degree C for the test actuator therefore an arbitrary continuous winding current must not result in exceeding the particular value, or otherwise, the actuator's internal processor would limit the PWM value to the BLDC motor, making any operation unpredictable. 

Experimental identification of the electro analogous circuit parameters is conducted with the data gathered for similar step input experiments with different open loop commanding PWM values. Steady state response for equations \ref{node one thermal equation}, \ref{node two thermal equation} and \ref{node three thermal equation} as well as model fitting for the experimental data resulted in an accuracy of 99.98 percentage in correctly interpreting the observed behaviour. Results for the model parameters are given in appendix Table \ref{table_parameters}.

Expected stator winding temperature at the steady state ($T_{W|s.s.}$) for a known continuous winding current $i_m$ can be calculated from the thermal model when the capacitors are saturated; thus open circuited.

\begin{equation}
	\label{fourth thermal equation}
	T_{W|s.s.}=T_A+\frac{i_m^2R_A(R_1+R_2+R_3)+iR_3}{1-\alpha_{cu}i_m^2R_A(R_1+R_2+R_3)}
\end{equation}

The nominal winding current $i_N$ is defined as the continuous stator winding current value that results in having the maximum permissible winding temperature reached in a steady state operation; i.e. $i_m=i_N$ when $T_{W|s.s.}=T_{MAX}$. According to the equation \ref{fourth thermal equation}, any decrease in $T_A$ would yield a higher value for $i_N$ and vice versa implying that $i_N$ depends on $T_A$ in practical applications.

When the actuator is undergoing some periodic operations, equation \ref{fourth thermal equation} will be governed by the root mean square (RMS) value of the stator winding current. In a practical application with multiple and identical actuators, the actuator with the highest RMS stator winding current or usually the actuator that contributes to the maximum joint torque value would thermodynamically limit the maximum capabilities of the entire system.  

It is possible to overload the internal BLDC motor beyond its nominal winding current for a brief period of time. This situation is characterised by the stator housing behaving as the ambiance to form a parallel $R_{1}$ and $C_{1}$ circuit subsection. The equation \ref{sixth thermal equation} assumes $T_\beta$ as the expected steady state stator winding temperature at an overloaded continuous current. In practice, as $T_\beta>T_{MAX}$, extended overloading would result in demagnetisations in the motor making it permanently unusable. Therefore, the overloading should be brief and another parameter should be defined for the maximum permissible time for a particular overloading condition. 

\begin{equation}
	\label{sixth thermal equation}
	T_W -T_H = (T_\beta-T_H)(1-e^{-t/\tau_1})
\end{equation}

A better intuition can be obtained in comparing the overloading current $i_O$ with the nominal winding current $i_N$. Equation \ref{seventh thermal equation} defines an overload constant `$K_o$' with two approximations made in the thermal model. Electrical stator winding resistance at $T_{MAX}$ winding temperature is considered equal to the stator winding resistance at $T_\beta$ winding temperature. Also the losses due to the solid state electronic heat flux $i$ are assumed to be negligible.

\begin{equation}
	\label{seventh thermal equation}
	K_o=\dfrac{i_O}{i_N}\sqrt{\dfrac{(T_{MAX}-T_A)R_1}{(T_{MAX}-T_H)(R_1+R_2+R_3)}}
\end{equation}

Therefore, the actuator's brief overloading can be characterised by the following equation that is now dependent on the defined overloading factor. $t_{on}$ is the maximum safe on-time to overload the actuator for a given $K_o$ value.

\begin{equation}
	\label{eighth thermal equation}
	t_{on}=\tau_1\,ln(\frac{{K_o}^2}{{K_o}^2-1})
\end{equation}

\begin{figure}[H]
	\centering
	\includegraphics[width=8.5cm]{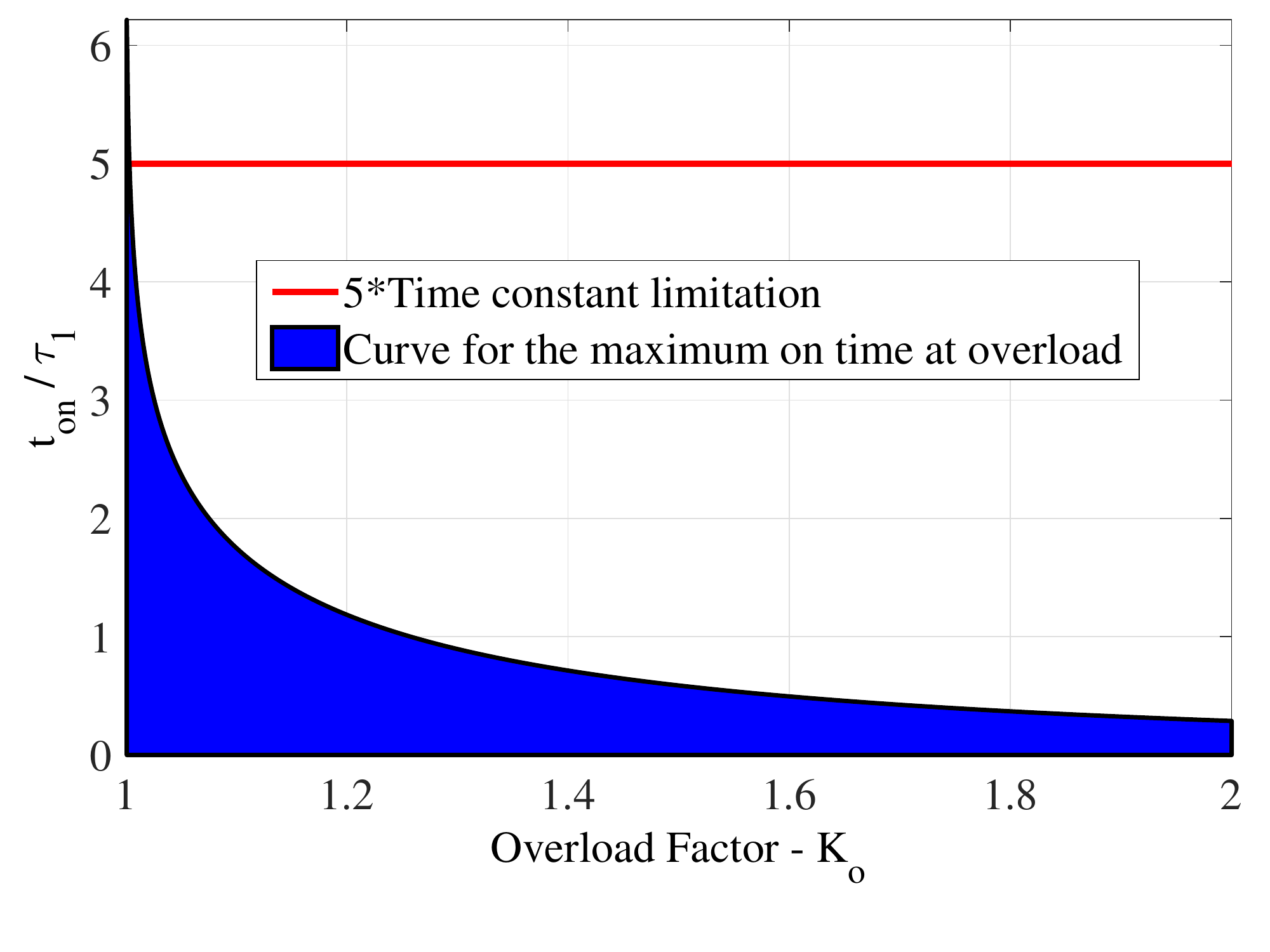} 
	\caption[Optional caption]{Maximum permissible on time.}
    \label{figure:overload}
\end{figure}

Figure \ref{figure:overload} represents the maximum allowable on time for a brief overloading session. The time is upper bounded to a five times of winding time constant (5$\times$1.492\,s) in order to be defined as a short term operation. It should be noted that the overloading factor depends on the initial housing temperature as well.

As the model parameters were calculated with respect to a heating up in the winding, in order to validate the results for a cooling down instance, the differential equations can be modified with a $U(-t)$ step function. For $t>0$, winding current is therefore zero resulting in no Joule power losses for the rest of the experimental duration. The governing time domain equations can be simulated and validated with respect to the actual experimental results.

\subsection{Stator winding temperature estimation}

Once the thermodynamic model is established from experimental results, it can be utilised along with the ambient temperature feedback and stator winding current feedback from the actuator to estimate the winding temperature inside the motor as in equations \ref{real time temp eqn01} and \ref{real time temp eqn01.1}. This approach is suitable to avoid the impracticality as well as the complexity of sensing the actual temperature inside any drive motor. Simplifying the thermodynamic model, it is assumed that the ambient temperature is equal to the temperature just outside the BLDC motor and $\tau_1<<\tau_2$.

\begin{equation}
	\label{real time temp eqn01}
	T_W(t)=T_H(t)- T_W(0)e^{\frac{-1}{R_1C_1}t}+\int_{0}^{t} P_{J}(\tau)g(t-\tau) d\tau
\end{equation}

\begin{equation}
	\label{real time temp eqn01.1}
	g(t)=\frac{1}{C_1}e^{-\frac{1}{R_1.C_1}t}
\end{equation}

In the equation \ref{real time temp eqn01}, $T_W(0)$ is the initial condition and $P_{J}(t)$ is the Joule losses due to the stator winding current as in the equation \ref{node zero thermal equation}. Instantaneous housing temperature can also be derived in a similar equation according to the simplified thermodynamic model.

\subsection{Model based controller design}
\label{dob section}

\begin{figure}[H]
	\centering
	\includegraphics[width=8cm]{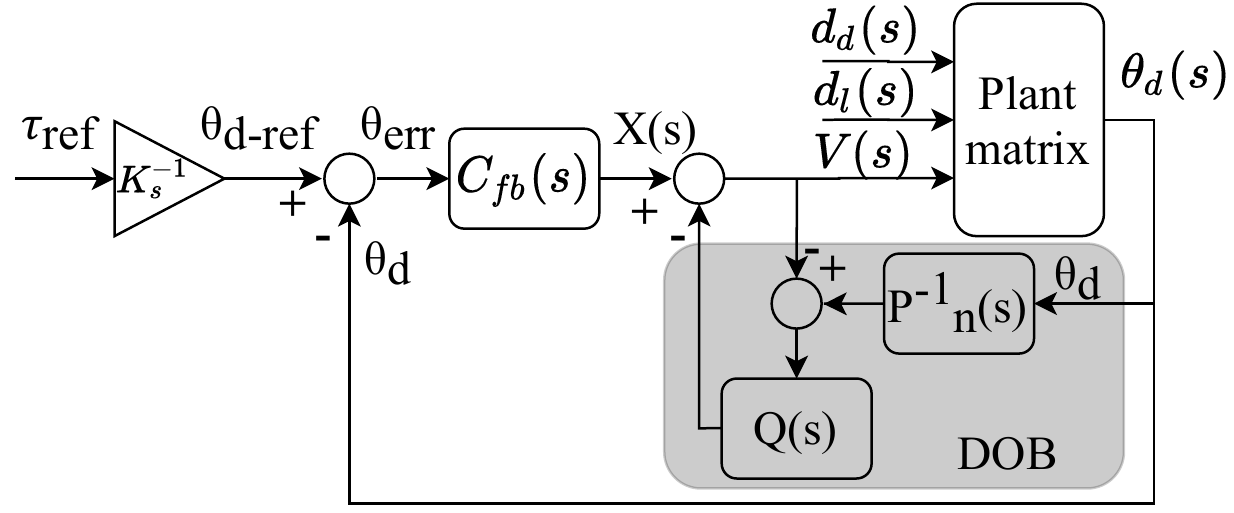} 
	\caption[Optional caption]{Proposed control system inclusive of the feedback controller and the disturbance rejection controller.}
    \label{figure:dob}
\end{figure}

The concept of the disturbance observer has been long utilised by researchers due to its performance in precise and robust control in force related applications. Following relationship in equation \ref{DOB_matrix_1} represents the input output behaviour of the system with reference to the terms given in Figure \ref{figure:SEAmodel} and Figure \ref{figure:dob}. $P_n(s)+\Delta(s)$ is the nominal plant model added with model uncertainties. Transfer functions $P_{d1}(s)$ and $P_{d2}(s)$ represent the contribution of external disturbances $d_d(s)$ and $d_l(s)$ into the observed output $\theta_d(s)$.

\begin{equation}
\label{DOB_matrix_1}
\begin{bmatrix}
     \theta_d(s)\\
\end{bmatrix}=
\begin{bmatrix}
     P_n(s) + \Delta(s)
     & P_{d1}(s)
     & P_{d2}(s) \\
\end{bmatrix}
\begin{bmatrix}
     V(s)\\
     d_d(s)\\
     d_l(s)\\
\end{bmatrix}
\end{equation}

The observer highlighted in grey in the Figure \ref{figure:dob} corresponds to an open loop DOB system where the compensation is applied only to the actuator plant excluding the feedback controller $C_{fb}(s)$. $X(s)$ is the output coming into the actuator from feedback controller. One of the most common observer type of nominal plant pseudo inversion is used here and the transfer function for the observer block can be deduced as;

\begin{multline}
	\label{DOB}
	\frac{\theta_d(s)}{X(s)} = \frac{P_n(s) +\Delta(s)}{1-Q(s)+ [P_n(s)+\Delta(s)][Q(s)P_n^{-1}(s)]}+\\
	[1-Q(s)]\frac{P_{d1}(s)d_d(s) + P_{d2}(s)d_l(s)}{X(s)}
\end{multline}

Here $Q(s)$ is a Butterworth low pass filter with a degree greater than or equal to the relative order of $P_n(s)$. Considering the behaviour of the system, at low frequencies, $Q(s)=1$ and results in;

\begin{equation}
    \label{DOB_resultant_eqn}
    \frac{\theta_d(s)}{X(s)} = P_n(s)
\end{equation}

 meaning the system will behave like the nominal plant model. For high frequencies, $Q(s)=0$ and the system would behave like the open loop plant with non modelled dynamics and disturbances.

$C_{fb}(s)$ is the feedback controller intended to compensate for the tracking error. It controls the reference of the pulse width modulated input to the actuator using a PID controller with suitable gains and clamps. MATLAB Simulink based PID tuning is conducted according to the response time and robustness required. Tuned controller augments the open loop cut off frequency of the actuator to a higher value and would additionally suppress the resonances as well. In the realisation, discrete time counterparts $C_{fb}(z)$, $Q(z)$ and $P^{-1}_n(z)$ are used according to the selected sampling rate. Inclusion of the Q filter is to stabilise the model inversion transfer function. In order to properly realise the filter, the Q filter must be stable alone and the magnitude should be less than the magnitude of the inverse of model-uncertainty upper bound. Also the cut off frequency should lie away from the interested torque bandwidth region.

\section{Experimental results}
\label{results}

Torque tracking of a SEA stands for the accurate control of the output torque to the load dynamics via the deflection of the elastic element. As only the stiffness of the elastic element is considered, it inherits static characteristics therefore, the scenario of torque control could be considered as a position control problem. Experimental actuator uses encoder readings to measure the deflection and return the torque value as a feedback. Within the scope of this section, torque tracking results in conjunction with thermodynamic behaviour of the actuator are discussed.

\subsection{Open loop torque response}

In open loop configuration, the fixed-output SEA is commanded with a sinusoidal reference PWM signal which linearly increases in the frequency. No feedback based control algorithm is enabled, therefore, the torque is effectively generated according to the open loop plant dynamics. Stator winding temperature behaviour is monitored and different experiments are conducted varying the amplitude of the reference signal.

\begin{figure}[ht]
	\centering
	\includegraphics[width=8.5cm]{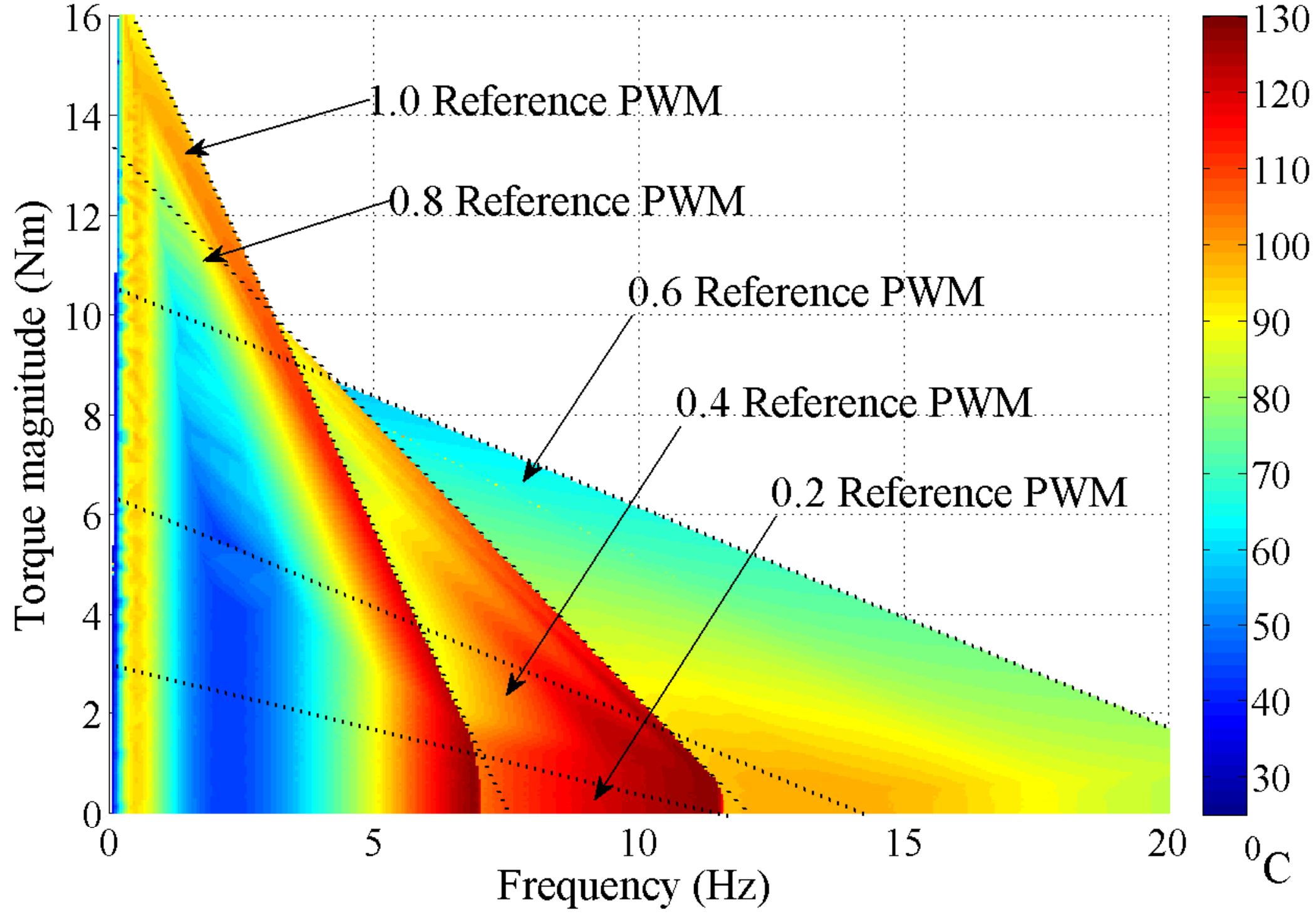} 
	\caption[Optional caption]{Torque magnitude variation with the frequency.}
    \label{figure:surf01_top}
\end{figure}

\begin{figure}[ht]
	\centering
	\includegraphics[width=8.5cm]{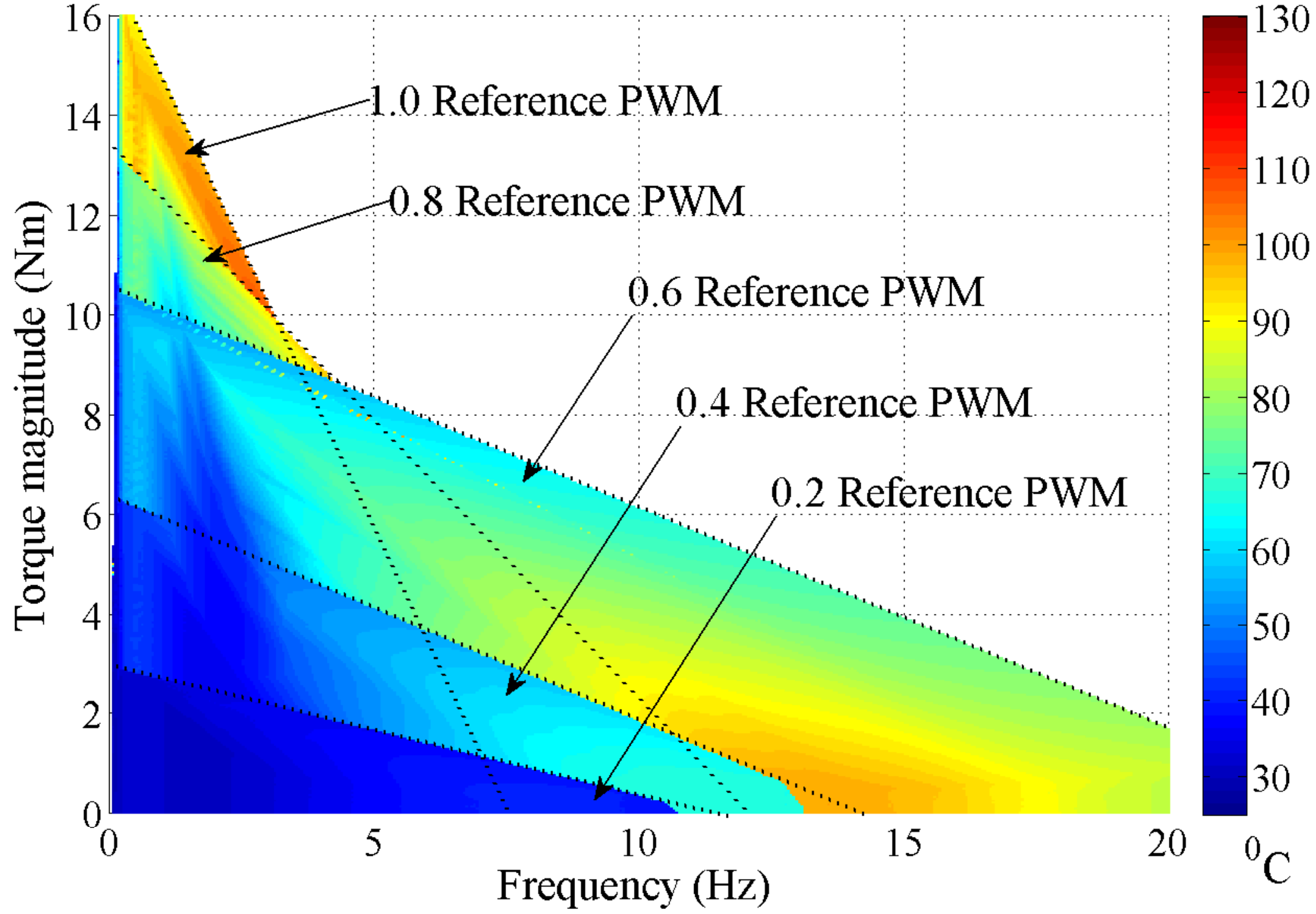} 
	\caption[Optional caption]{Torque magnitude variation graph emphasising the lower reference values.}
    \label{figure:surf01_bottom}
\end{figure}

In Figure \ref{figure:surf01_top}, time domain behaviour is omitted and instead, the torque magnitude variation at each instantaneous frequency is plotted along with the instantaneous winding temperature value. Figure contains the actual data points as well as some interpolated data points for better visualisation. The reference signal is duly terminated if the winding temperature is reached $T_{MAX}$, the maximum permissible value of 130\degree C and no more data points are acquired for higher frequency values. As the winding temperature values relatively become higher with higher reference amplitudes, behaviours corresponding to 0.4 and 0.2 reference amplitudes are not visible in the figure and are emphasised in the figure \ref{figure:surf01_bottom}.

The results from figures \ref{figure:surf01_top} and \ref{figure:surf01_bottom} emphasise the torque generation characteristics of the actuator with respect to the frequency of the reference signal. From the respective transfer function analysis, minimum open loop torque bandwidth is obtained as 1.54\,Hz and no thermal limitations are observed before the particular value. The surfaces are not plotted beyond 20\,Hz. 

A noticeable trench that portrays lower winding temperatures can be observed in the figure \ref{figure:surf01_top}. Initially at the lower frequencies, the actuator can generate torque values with a higher gain, therefore, generates more Joule losses and it later decreases as the torque gain reduces with the frequency but eventually reaches higher Joule losses again as the frequency increases.

Output locked scenario is optimal for benchmarking instances but in reality the torque utilisation is much more relevant for the output non-locked scenarios. It corresponds to having a non-infinite load side inertia ($J_l$).  

\begin{figure}[ht]
\centering 
    \includegraphics[width=8.5cm]{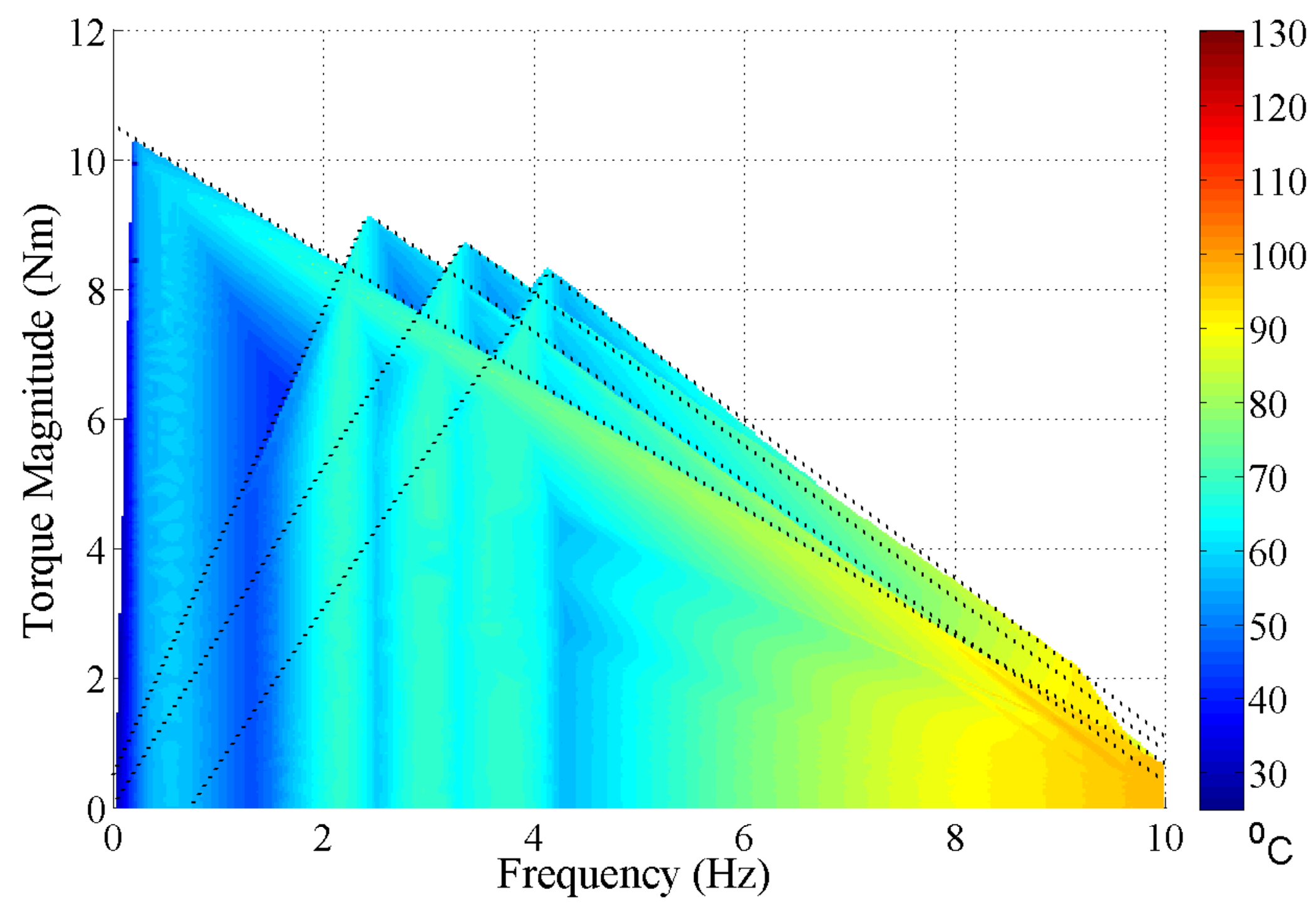} 
    \caption{Torque magnitude variation for non-zero load side dynamics.}
    \label{figure:eff v inertia}
\end{figure}

Figure \ref{figure:eff v inertia} shows the torque characteristics for four different load side dynamics and the open loop input signal is a 0.6 reference PWM signal. The frequency where the highest torque magnitudes are produced, drifts from left to right when the load dynamics given by $P_l(s)$ starts from $0$ (output fixed) and increased (load side inertia $J_l$ is reduced). According to the winding temperature based colour mapping in the figure, it can be concluded that the highest winding temperatures are predominantly obtained from the output fixed instance during higher frequencies, therefore, the possible thermal limitations would be visible there first. 

\subsection{Closed loop torque tracking}

The torque tracking capabilities in a closed loop system will be governed by the feedback controller algorithm and is subjective to change with alterations in the controller parameters. For example, re-tuning proportional, integral and derivative gains of a PID controller will change the response of a system to a given input. In order to provide a good estimation on the accessible torque bandwidth of the actuator, model based control algorithm parameters are obtained using Simulink PID tuning.

\begin{figure}[ht]
      \centering
      \includegraphics[width=8.5cm]{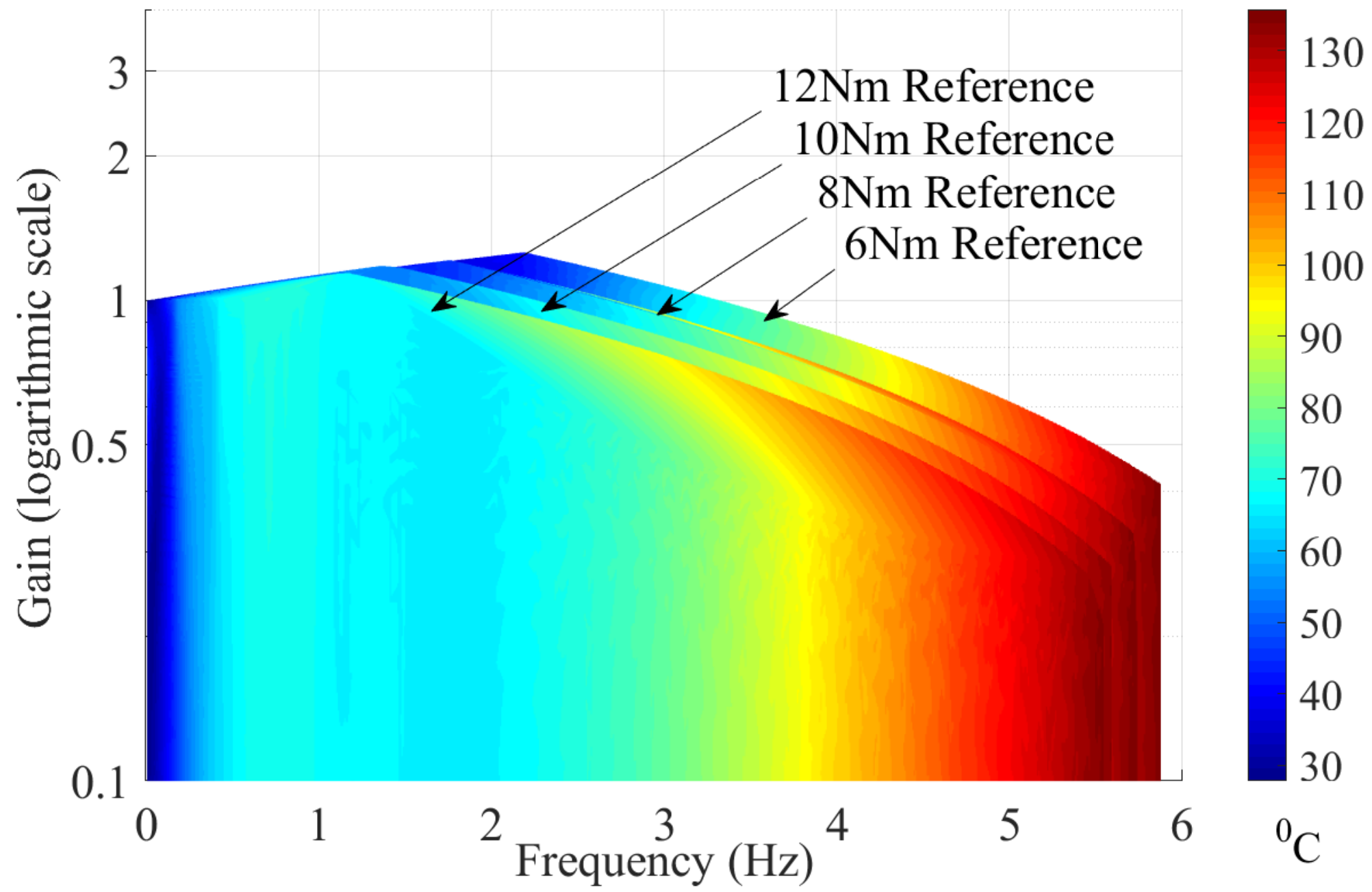} 
      \caption{Closed loop output torque gain variation with frequency.}
      \label{figure:closed loop surface}
\end{figure}

Figure \ref{figure:closed loop surface} visualises the torque tracking results for the closed loop system superimposed with the stator winding temperature variation throughout the frequency scale. Maximum permissible value of 130\degree C is reached around 5\,Hz implying a possible thermal limitation occurrence in prior to the reaching of the closed loop torque bandwidth.

\begin{figure}[ht]
      \centering
      \includegraphics[width=8.5cm]{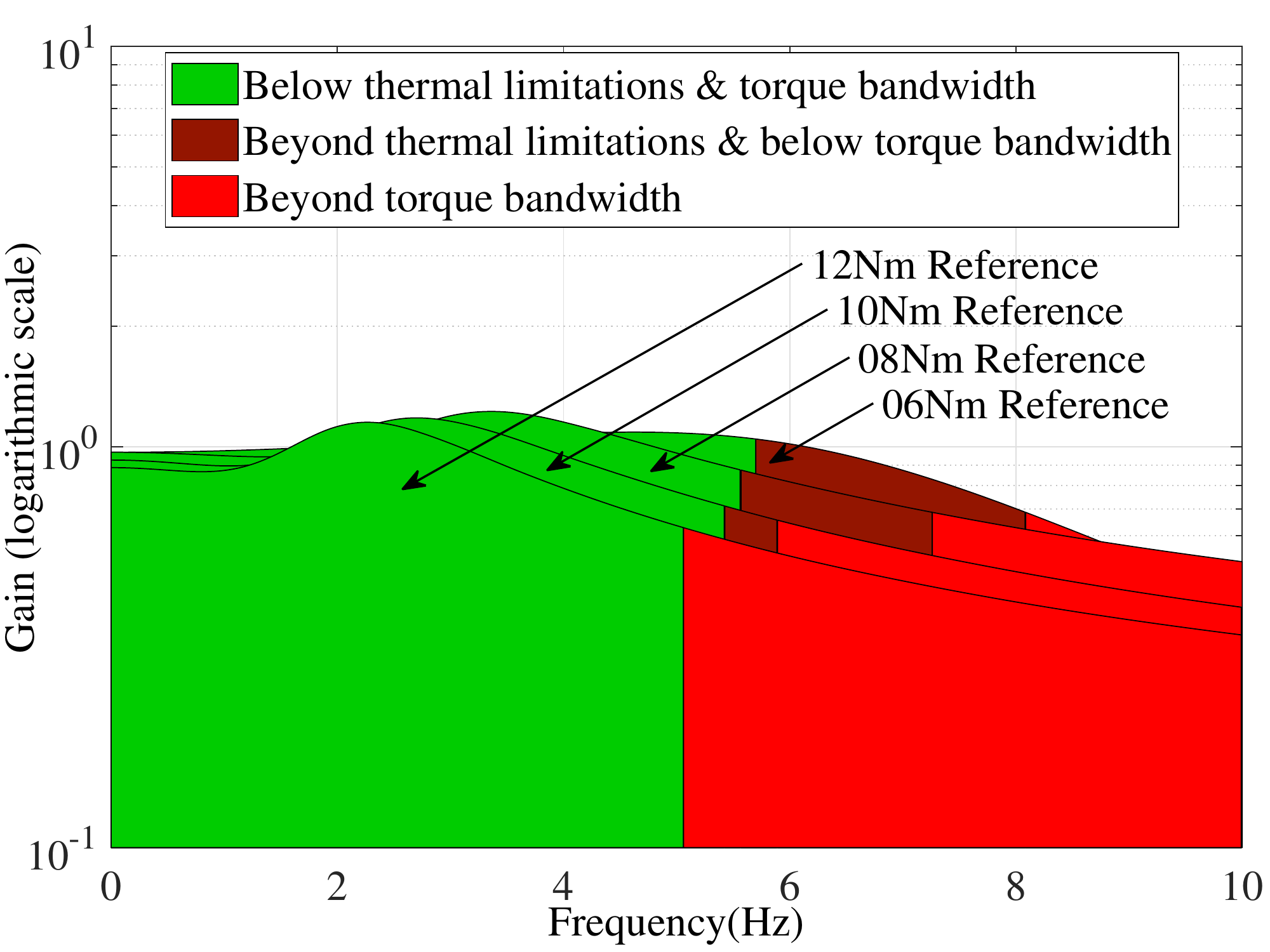} 
      \caption{Comparison of the closed loop torque bandwidth and the frequency at thermal limitations.}
      \label{figure:bandwidth compare}
\end{figure}

The Figure \ref{figure:bandwidth compare} shows the gain transfer functions derived for each input reference torque amplitude shown in Figure \ref{figure:closed loop surface}. For 6, 8 and 10\,Nm reference values, the actuator could not reach the characteristic closed loop torque bandwidth due to the occurrence of thermal extremes in a prior frequency value. The actuator could reach the torque bandwidth prior to the thermal extreme frequency only in the 12\,Nm amplitude reference signal. Unlike in the open loop system, the general thermal behaviour for each of the reference signals remains the same in closed loop configuration as the feedback controller drives the PWM reference command signal to the saturation regions (+1 and -1) after a certain frequency value.  

Occurrence of thermal limitations prior to the closed loop torque bandwidth imposes a disadvantage over the wide range of usability of an actuator. Nevertheless, considering only the closed loop electromechanical bandwidth of the actuator while excluding the thermodynamics could result in a shorter lifespan of the actuator when operated in thermal extreme regions. Following section discusses on the inclusion of a thermal controller which would regulate the inputs to the actuator.

\subsection{Regulation of PWM reference commands to avoid thermal limitations}
\label{Adaptive regulation example}

According to the experimental results, the expected torque bandwidth is not accessible due to thermal limitations, therefore, a controller should regulate the input PWM reference signal beyond a certain winding temperature threshold. A predefined amplitude restriction as well as low pass filtering would ideally delay the thermal extremes but as it alters the input signal waveform, would consequently lower the torque bandwidth as well. However the trade-off is justifiable in order to harness the maximum accessible torque bandwidth. 

\begin{figure}[ht]
	\centering
	\includegraphics[width=8.5cm]{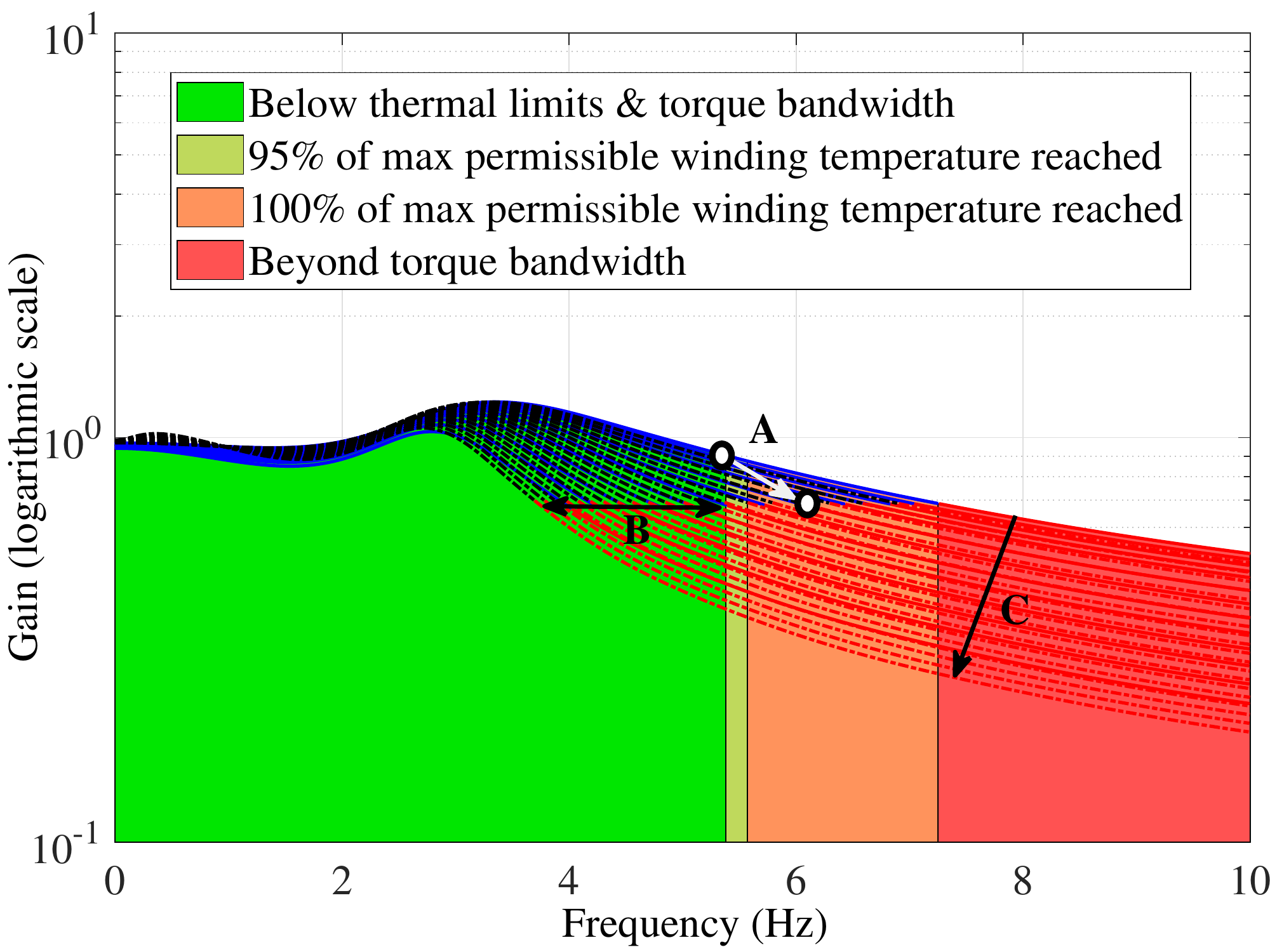}
	\caption[Optional caption]{Example thermal control for the torque amplitude of 8\,Nm. Solid lines represent a transfer function reshape from gain adjustment and dashed lines represent a transfer function reshape from non linear filtering.}
    \label{figure:thermal controller}
\end{figure} 

The thermal controller is triggered at the threshold value of $95\%$ of the maximum permissible winding temperature $T_{MAX}$ (Starting point of arrow `A'). As the temperature increases beyond the value, the controller lowers the PWM input reference signal's gain and removes some of the high frequency components. Consequently, the input-output transfer function would also deviate from the original transfer function as shown from the arrow `C' of the figure \ref{figure:thermal controller}. An example scenario is emphasised from the arrow `A'. The end point of the arrow corresponds to the refined closed loop torque bandwidth with no thermal limitations and such a bandwidth improvement is significant in comparison to the interested frequency range. Transfer function reshaping by the controller within the range given by the arrow `B' will have a closed loop bandwidth less than the original accessible value and the remaining region will have an improved accessible torque bandwidth. 

\section{Discussion and conclusions}
\label{conclusion}

Torque tracking capabilities of an actuator depend on the characteristics of the electromechanical subsystem, thermodynamic subsystem as well as on the implemented controller algorithms, therefore, the limitations could occur with any or all of them. Within the scope of the paper, the detailed methodology in evaluating the accessible torque bandwidth value of an actuator is emphasised while taking all such subsystems into the parallel consideration.

Experimental results show the presence of thermodynamic limitations near the electromechanical torque bandwidth. The experiments were conducted in a practical scenario of closed loop configuration and it was identified that the output fixed instance will capture the thermodynamic limitations first in the frequency range. The authors have further evaluated an approach in enhancing the accessible torque bandwidth with incorporation of a thermal controller. In the example shown in subsection \ref{Adaptive regulation example}, the electromechanical torque bandwidth of 7.25\,Hz was not accessible due to the stator winding temperature exceeding the maximum permissible value at 5.57\,Hz. Inclusion of the thermal controller improves the accessible torque bandwidth by $8.25\%$ up to 6.03\,Hz. Authors therefore encourage the readers in inclusion of such an analysis for maximising the accessible torque bandwidth as well as the lifespan of actuators in torque control applications. 
Opportunities for further research are identified and should prioritise in evaluating the accessible torque bandwidth with respect to the implemented controller characteristics such as load resonances. Within the scope, only feedback and disturbance rejection based control systems are utilised and future work can include more robust algorithms such as inclusion of a feedforward control layer to the system. External input to the electrical subsystem is considered to be only the PWM reference signal, therefore, the dependency of the stator winding current is considered to be a multiple input function of the reference PWM value and of the instantaneous motor velocity. However, if the thermal controller is capable enough to regulate the winding current as well, the transfer functions and control systems derived within the paper's scope should be revised accordingly.

\section{Acknowledgement}

The authors would like to thank Arash Kh. Sichani for his guidance and advice during this work. Authors would also like to thank Isuru Kalhara and Thomas Molnar for their support. This work was fully funded by Commonwealth Scientific and Industrial Research Organisation (CSIRO), Australia.

\section*{Appendix}

The actuator under experimental consideration is a HEBI robotics professional grade series elastic actuator named X5-9. Electromechanical model parameters are given in Table \ref{table_parameters_2} and thermodynamic model parameters are given in Table \ref{table_parameters}.

\begin{table}[H]
  \caption{Four body thermodynamic model parameters of the actuator}
  \label{table_parameters}
  \begin{center}
  \begin{tabular}{rl}
  \toprule
    Parameter & Value\\
  \midrule
    Thermal time constant $\tau_{1}$ & $1.49\,s$\\
    Thermal time constant $\tau_{2}$ & $13.66\,s$\\
    Thermal time constant $\tau_{3}$ & $>60\,s$\\
    Thermal resistance $R_{1}$ & $5.368\,KW^{-1}$\\
    Thermal resistance $R_{2}$ & $1.253\,KW^{-1}$\\
    Thermal resistance $R_{3}$ & $0.357\,KW^{-1}$\\
  \bottomrule
  \end{tabular}
  \end{center}
\end{table}

\begin{table}[H]
  \caption{Electromechanical model parameters of the actuator} 
  \label{table_parameters_2}
  \begin{center}
  \begin{tabular}{rl}
  \toprule
  Parameter & Value \\
  \midrule
  Gear ratio $N:1$ & $1742.222:1$ \\
  
  Spring constant $K_s$ & $130\,Nmrad^{-1}$ \\
  
  Torque constant & $7.1\,NmA^{-1}$ (actuator) \\
  
  Speed constant & $0.9\,rpm V^{-1}$ (actuator) \\
  
  Terminal resistance & $9.99\,\Omega$ (phase to phase)\\
  
  Terminal inductance  & $0.163\,mH$ (phase to phase)\\
  
  Torque constant & $6.26\,mNmA^{-1}$ (BLDC) \\
  
  Speed constant & $1530\,rpm V^{-1}$ (BLDC) \\
  
  Experimental $J_m $  & $5.615*10^{-8}\,kgm^{2}$  \\
  
  Experimental $B_m $  & $8.726*10^{-7}\,Nms{rad}^{-1}$ \\
  
  Experimental $K_{\tau} $  & $5.484*10^{-3}\,NmA^{-1}$ \\
  
  Experimental $K_e $  & $2.8*10^{-3}\,Vs{rad}^{-1}$ \\
  
  Experimental L & $0.794\,mH$ \\
  
  Experimental R & $6.840\,\Omega$\\
  \bottomrule
  \end{tabular}
  \end{center}
\end{table}

\balance
\bibliographystyle{named} 

\end{document}